\documentclass[lettersize,journal]{IEEEtran}
\usepackage{amsmath,amsfonts}
\usepackage{array}
\usepackage{textcomp}
\usepackage{subfigure}
\usepackage{url}
\usepackage{verbatim}
\usepackage{graphicx}
\usepackage{multirow}
\usepackage{booktabs}
\usepackage{subfloat}
\usepackage{float}
\usepackage{cite}
\usepackage{color}
\usepackage{textcase}
\usepackage{algorithm}
\usepackage{amsfonts,amssymb} 
\usepackage{amsmath}
\usepackage{multicol}
\usepackage{stfloats}
\usepackage{graphicx}
\usepackage{amsmath}
\usepackage{amssymb}
\usepackage{booktabs}
\usepackage{multirow}
\usepackage{caption}
\usepackage{graphicx}
\usepackage{float} 
\usepackage[table,xcdraw]{xcolor}
\usepackage{enumerate}
\usepackage{enumitem}
\usepackage{amsthm,amsmath,amssymb,mathrsfs}
\usepackage{booktabs}
\usepackage{listings}
\usepackage{threeparttable}
\def\BibTeX{{\rm B\kern-.05em{\sc i\kern-.025em b}\kern-.08em
    T\kern-.1667em\lower.7ex\hbox{E}\kern-.125emX}}
\usepackage{balance}
\usepackage[colorlinks,bookmarks=false,citecolor=green]{hyperref}
\usepackage{algpseudocode}

\begin{document}
\title{Learning Mutual Excitation for Hand-to-Hand and Human-to-Human Interaction Recognition}

\author{
Mengyuan Liu, Chen Chen, Songtao Wu, Fanyang Meng, Hong Liu$^{\dagger}$

\thanks{M. Liu, H. Liu are with State Key Laboratory of General Artificial Intelligence, Peking University, Shenzhen Graduate School.
Chen Chen is with OPPO AI Center.
Songtao Wu is with Sony R\&D Center China.
Fanyang Meng is with Peng Cheng Laboratory.
This work was supported by Shenzhen Innovation in Science and Technology Foundation for The Excellent Youth Scholars (No. RCYX20231211090248064). Corresponding author: Hong Liu (hongliu@pku.edu.cn).}
}

\maketitle

\begin{abstract}
Recognizing interactive actions, including hand-to-hand interaction and human-to-human interaction, has attracted increasing attention for various applications in the field of video analysis and human-robot interaction. Considering the success of graph convolution in modeling topology-aware features from skeleton data, recent methods commonly operate graph convolution on separate entities and use late fusion for interactive action recognition, which can barely model the mutual semantic relationships between pairwise entities. To this end, we propose a mutual excitation graph convolutional network (me-GCN) by stacking mutual excitation graph convolution (me-GC) layers. Specifically, me-GC uses a mutual topology excitation module to firstly extract adjacency matrices from individual entities and then adaptively model the mutual constraints between them. Moreover, me-GC extends the above idea and further uses a mutual feature excitation module to extract and merge deep features from pairwise entities. Compared with graph convolution, our proposed me-GC gradually learns mutual information in each layer and each stage of graph convolution operations. Extensive experiments on a challenging hand-to-hand interaction dataset, i.e., the Assembely101 dataset, and two large-scale human-to-human interaction datasets, i.e., NTU60-Interaction and NTU120-Interaction consistently verify the superiority of our proposed method, which outperforms the state-of-the-art GCN-based and Transformer-based methods. Code is provided in https://github.com/nkliuyifang/me-GCN.
\end{abstract}



\section{Introduction}
Interactive action recognition is an important task, as it is an indispensable stream of action recognition \cite{naser2022privacy, wang2022predicting, hadikhani2023novel, taghanaki2023self, baruah2023intent} that has applications in video analysis \cite{liu2022generalized, liu2023temporal, wang2024dynamic, zhang2024facial, wang2023global, tu2023dtcm, tu2023consistent, gao2023dual} and human-robot interaction \cite{zhang2016rgb, ahmad2021graph, pareek2021survey, kong2022human}. 
Recently, skeleton-based interactive action recognition has attracted increasing attention for the development of depth sensors, e.g., Kinect \cite{zhang2012microsoft} and Intel RealSense \cite{keselman2017intel}, and the robustness of skeleton data to complicated backgrounds and viewpoint variations \cite{liu2017enhanced, sun2022human}.
We focus on interaction between \textbf{\textit{pairwise entities}}, including hand-to-hand and human-to-human interaction.

\begin{figure}[t]
  \centering
   \includegraphics[width=0.95\linewidth]{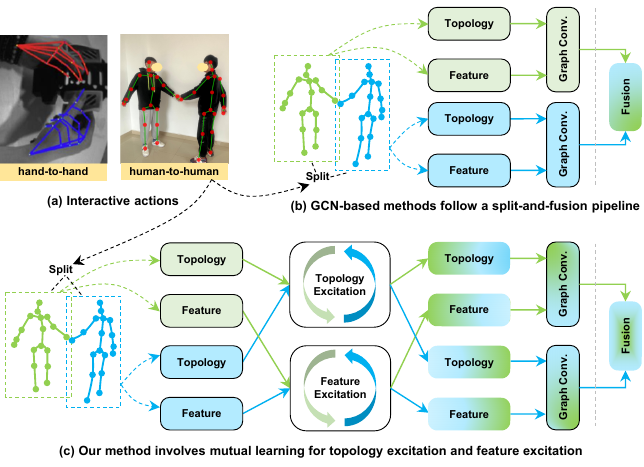}
   \vspace{-0.5em}
   \caption{\textbf{Illustration of our general idea}. To recognize interactive actions, e.g., ``high five" and ``handshaking" (a), previous GCN-based methods follow a split-and-fusion pipeline (b), which overlooks the mutual semantic relationships between interactive body parts. To solve this problem, our method involves mutual learning for topology excitation and feature excitation (c). Note that only one layer is shown in (b) and (c) for simplicity.}
   \label{fig1}
   \vspace{-1.5em}
\end{figure}

Pioneering deep-learning-based methods treat human joints as a set of independent features \cite{perez2021interaction} or aggregate manually-defined human joints as a set of local features \cite{pang2022igformer}, which are fed into Recurrent Neural Network (RNN) \cite{schuster1997bidirectional} or Transformer \cite{vaswani2017attention} to predict interactive action labels. Despite the effectiveness of encoding local features, these methods ignore the human body topology that reveals inherent global correlations between joints. As a human skeleton is a natural graph, recent methods \cite{yan2018spatial, cheng2020decoupling} present graph convolutional network (GCN) to learn human body topology-aware features from a human skeleton, where each graph convolution (GC) layer uses a convolution to extract local feature and further uses a topology modeling module to aggregate local feature into the global feature. To encode pairwise skeletons in an interactive action, these methods adopt a \textbf{\textit{split-and-fusion}} pipeline (see Fig. \ref{fig1} (b)), where an interactive action is firstly split into two skeleton sequences, and then each sequence is encoded by multiple graph convolution layers and finally extracted features for both sequences are merged by late fusion.

However, the split-and-fusion pipeline overlooks the mutual semantic relationships between interactive body parts, which serve as critical cues to describe an interactive action.
Taking two similar interactive actions called ``high five" and ``handshaking" as an example (see Fig. \ref{fig1} (a)), the interactive hands of two persons are the key components to distinguish them. 
Motivated by this observation, we ask a natural question: \textbf{\textit{How to enable GCN-based methods for encoding mutual semantic relationships between interactive body parts?}}
As a first attempt, we answer this question by injecting mutual learning into the topology modeling module of the standard graph convolution operation. Our motivations are two-fold. First, topology modeling severely affects the performances of graph convolution \cite{li2019actional}. Second, the correlations between joints for each human skeleton depend on both human skeletons. Therefore, we present a mutual topology excitation module (MTE) for topology learning from interactive human skeletons.
As the performance of the graph convolution operation also depends on the extracted local features, we extend the idea of MTE and present a mutual feature excitation module (MFE) for local feature learning. We infer that mutual information serves as attention to adjust the importance of local features.

Taking advantage of MTE and MFE, we present \textbf{\textit{a mutual excitation graph convolution (me-GC) to involve mutual information exchange in both topology modeling and feature extraction stages}}. In Fig. \ref{fig1} (c), we show the general idea of one me-GC layer for simplicity. By stacking multi-layer me-GC for extracting cascade features, we build a mutual excitation graph convolutional network (me-GCN). The pipeline of me-GCN is depicted in Fig. \ref{fig2}, where each me-GC layer consists of an MTE module and an MFE module. Generally, the proposed cascade me-GC layers enable graph convolution operations to effectively encode the mutual semantic relationships between interactive body parts. Our main contributions are three-fold:

\begin{itemize}
\item Compared with GCN-based methods that follow a split-and-fusion pipeline for skeleton-based interactive action recognition, we leverage mutual learning to enable GCN-based architectures for efficient encoding of mutual semantic relationships between interactive body parts.

\item Beyond standard GC, we present a mutual excitation graph convolution (me-GC) for topology-aware feature extraction using mutual learning, where a mutual topology excitation module (MTE) is designed for mutual topology learning and a mutual feature excitation module (MFE) is developed for mutual local feature learning.

\item We present a mutual excitation graph convolutional network (me-GCN) for skeleton-based interactive action recognition. Compared with both GCN-based and Transformer-based methods, we achieve superior performances on hand-to-hand interaction recognition task and human-to-human interaction recognition task.
\end{itemize}

\section{Related Work}
Skeleton-based interactive action recognition can be achieved by either designing specific models for this task or generalizing existing skeleton-based action recognition models for this task through the split-and-fusion pipeline.

\textbf{Specific Models.}
Researchers have developed specific models for skeleton-based interactive action recognition task by exploring the mutual relationships using hand-crafted-based \cite{zhang2012spatio, ji2014interactive, ji2015learning, wu2017recognition}, RNN-based \cite{perez2021interaction} and Transformer-based \cite{pang2022igformer} models.
Hand-crafted models either focus on mining interactive body parts using spatial-temporal phrases \cite{zhang2012spatio}, interactive body part contrast mining \cite{ji2014interactive} and contrastive feature distribution model \cite{ji2015learning}, or focus on selecting representative joint features \cite{wu2017recognition}.
Incorporating RNN architecture, an interaction relational network \cite{perez2021interaction} has been proposed to reason over the relationships between skeleton joints during interactions.
Recently, an interaction relational network \cite{pang2022igformer} develops a parallel interactive transformer, which takes advantage of the transformer to reason over relationships between skeleton joints during interactions.
Generally, recent specific methods for skeleton-based interactive action recognition mainly apply RNN and Transformer to model the spatial-temporal relationships among local features, which ignores encoding human body topology therefore finds difficulties in encoding global mutual semantic features.  

\textbf{Generalized Models.}
Existing skeleton-based action recognition methods can be generalized to recognize skeleton-based interactive actions with the split-and-fusion pipeline. 
We review the main-stream skeleton-based action recognition methods including using RNN, CNN, and GCN. As a pioneering work, the hierarchical RNN \cite{du2015hierarchical} was the first work to apply RNN for modeling the long-term contextual information of skeleton sequences. Based on RNN, a spatial-temporal LSTM \cite{liu2016spatio} was presented to encode hidden information of the skeleton over both spatial and temporal domains. To enhance the sequential modeling ability of RNN, a spatial-temporal transformer network \cite{plizzari2021skeleton} was used to understand intra-frame interactions between different body parts and to model inter-frame correlations. Instead of using a sequential modeling network, a CNN model \cite{du2015skeleton} was used to model the hidden spatial-temporal information of skeleton sequences from an image, which is the concatenation of the joint coordinates. Moreover, multiple images \cite{ke2017new, liu2017enhanced} were also used as inputs of CNN models to extract spatial-temporal skeleton features. Recently, a Spatial-Temporal Graph Convolutional Network (ST-GCN) \cite{yan2018spatial} was proposed to model dynamic skeletons, which moves beyond the limitations of previous methods by automatically learning spatial and temporal patterns. 
Instead of using manually-defined adjacency matrices, neural searching was used for learning adjacency matrices \cite{peng2020learning}. Meanwhile, 2s-AGCN \cite{shi2019two} and AS-GCN \cite{li2019actional} were presented for data-dependant dynamic topology learning, which can create action-specific joint connections in the adjacency matrices.
Meanwhile, for the skeleton-based interactive action recognition task, we notice that the specific design for GCN-based methods is lacking, and directly using the split-and-fusion pipeline to generalize these GCN-based methods will inevitably fail to encode the mutual relationships among interactive entities.

\section{Proposed Method}
To learn mutual entity relationships in the interactive actions, we present a mutual excitation graph convolutional network (me-GCN).
This section introduces our me-GCN in detail.
First, we describe an overview of me-GCN.
Second, we generally define the input layer, the me-GC layer, the temporal convolution layer, and the inference layer of me-GCN.
Third, we specifically define the mutual topology excitation (MTE) module and the mutual feature excitation (MFE) module for implementing the me-GC layer. Noting that the feature generation block (FGB) and feature fusion block (FFB) are basic blocks to implement MTE and MFE. 

\subsection{Overview}\label{3-1}
\begin{figure*}[t]
  \centering
   \includegraphics[width=1.0\linewidth]{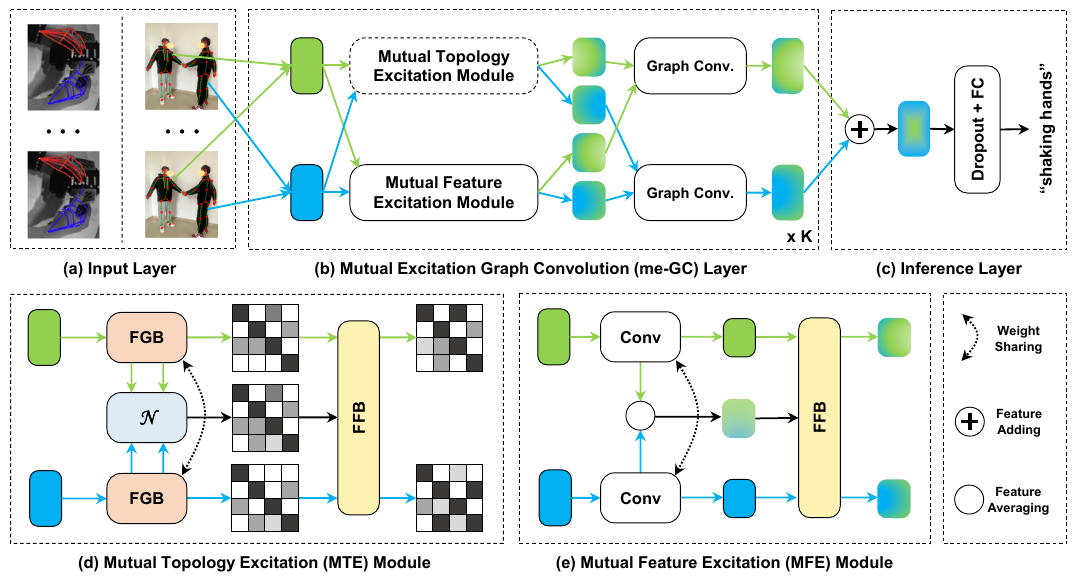}
   \vspace{-1em}
   \caption{\textbf{Overview of our proposed mutual excitation graph convolutional network (me-GCN)}, which contains an input layer, $K$ mutual excitation graph convolution (me-GC) layers, and an inference layer. Each me-GC layer contains a mutual topology excitation module (MTE), a mutual feature excitation module (MFE), and two graph convolution operations. FGB and FFB denote feature generation block and feature fusion block respectively, and function $\mathcal{N}$ (see Eq. \ref{eq4}) is used to fuse outputs from FGB.}
   \label{fig2}
   \vspace{-1em}
\end{figure*}

Fig. \ref{fig2} shows our me-GCN, which has three main steps.
First, we need to convert skeleton frames as feature vectors.
Specifically, we use the input layer to arrange skeleton frames as a feature vector which is normalized across different channels.
Then, we extract deep features with $K$ me-GC layers.
Beyond standard GC in previous approaches \cite{li2019actional, chen2021channel, chi2022infogcn}, our me-GC jointly learns mutual topology information using the MTE module and learns mutual feature information using the MFE module.
Inside of an MTE module, we introduce FGB to extract non-shared multi-channel adjacency matrices from each entity and use function $\mathcal{N}$ to extract shared multi-channel adjacency matrices from both entities and subsequently design FFB to adaptively fuse non-shared and shared multi-channel adjacency matrices.
Inside an MFE module, we use convolution operations to extract non-shared multi-channel deep features from each entity and use the feature averaging operation to extract shared multi-channel deep features from both entities, and subsequently use FFB to adaptively fuse non-shared and shared multi-channel deep features.
Finally, we integrate entity features of two paths with an inference layer to predict the interactive action label.

\subsection{me-GCN}\label{3-2}

\textbf{Input Layer}. A skeleton sequence $\textbf{S}_0 \in \mathbb{R}^{2\times C_0\times T_0\times N}$ denotes an interactive action with two entities, which contains $C_0$ channels, $T_0$ frames and $N$ joints for each skeleton. We normalize $\textbf{S}_0$ across channels, and the normalized data is denoted as $\textbf{X}_0 \in \mathbb{R}^{2\times C_0\times T_0\times N}$, which suffers less from abnormal coordinate values and benefits the convergence of deep neural networks. Specifically, we use a batch of skeleton sequences as the input in the training phase. Suppose the input data lies in $\mathbb{R}^{B\times 2\times C_0\times T_0\times N}$, where $B$ is the batch size. We map the input to $\mathbb{R}^{(B\times 2\times C_0\times N)\times T_0}$ and then normalize the data along the temporal axis. In the test phase, $\textbf{S}_0$ is firstly mapped to $\mathbb{R}^{(2\times C_0\times N)\times T_0}$, and then normalized along the temporal axis.

\textbf{me-GC}. 
After normalization, we use $K$ graph convolution layers to extract semantic features from the normalized skeleton sequence $\textbf{X}_0$. Graph convolution contains feature extraction and topology modeling. Considering the $k$-$th$ GC layer, we use the MFE module for feature extraction, which maps $\textbf{X}_k$ to $\textbf{F}_{k+1} \in \mathbb{R}^{2\times C_{k+1}\times T_{k}\times N}$.
For topology modeling, we use the MTE module to formulate the adjacency matrix $\textbf{{A}}_{k+1} \in \mathbb{R}^{2\times C_{k+1}\times N\times N}$ from $\textbf{X}_k$. Then, our me-GC layer is implemented as:
\vspace{-0.8em}
\begin{equation}
    \resizebox{0.88\linewidth}{!}{
    $\tilde{\textbf{X}}_{k+1} = \mathscr{G}\big(\textbf{F}_{k+1}, \textbf{A}_{k+1}\big) = \sum\limits_{d}{\textbf{F}_{k+1}[a,b,c,d]\cdot\textbf{A}_{k+1}[a,b,d,e]}$,}\vspace{-0.8em}
\end{equation}
where $\mathscr{G}\big(\textbf{$\cdot$}\big)$ function is implemented by $einsum$ defined in Pytorch \cite{paszke2019pytorch}, the extracted feature $\tilde{\textbf{X}}_{k+1}$ fuses deep features of neighborhoods defined by the adjacency matrix.

\textbf{Temporal Convolution (TC)}. Although the spatial relationships of pairwise skeletons can be encoded by graph convolution, we still lack in capturing the temporal relationships among skeletons located at different frames. To encode the temporal information, we use temporal convolution after each me-GC operation to capture short-term temporal relationships, which is formulated as:
\vspace{-0.5em}
\begin{equation}
    \textbf{X}_{k+1} = \mathscr{T}\big(\tilde{\textbf{X}}_{k+1}\big) = \sum_{i=1}^{I}{conv_i\big(\tilde{\textbf{X}}_{k+1}\big)},\vspace{-0.5em}
\end{equation}
where $I$ number of temporal convolutions are applied on $\tilde{\textbf{X}}_{k+1}$, and these generated features are fused to obtain the output feature $\textbf{X}_{k+1}$, which fuses both spatial and temporal features. Considering that a single skeleton contains noise that affects the robustness of the feature, we take advantage of temporal surrounding skeletons and use temporal convolution to smooth the underline noise in features, thus enhancing the robustness of features.

\textbf{Inference Layer}. 
After applying $K$ me-GC layers and TC operations, we obtain $\textbf{X}_K \in \mathbb{R}^{2\times C_K\times T_K\times N}$, which denotes the individual embedding of two skeleton sequences. 
We use late fusion to attend to the mutual information between two entities by summarizing the first dimension of $\textbf{X}_K$ to formulate a fused embedding, which is further passed through a dropout layer \cite{srivastava2014dropout} and a fully connected layer for the final prediction of interactive action labels.

\subsection{MTE \& MFE}\label{3-3}

As a core component of graph convolution, the adjacency matrix describes the relationships between graph nodes and guides the path of message passing \cite{zhang2019graph}. How to construct an adjacency matrix remains an open problem, especially for the skeleton-based action recognition task. Instead of using a single adjacency matrix, multiple high-order adjacency matrices can be jointly used in an inception module \cite{huang2020spatio}. Rather than design a hand-crafted adjacency matrix, a dynamic GCN is designed to use a data-driven adjacency matrix that can be optimized through backward propagation \cite{shi2019two, ye2020dynamic}. Inspired by CNN which uses an independent spatial aggregation kernel for every channel to capture different spatial information, the decoupling GCN adopts an independent adjacency matrix for every channel to boost the graph modeling ability with no extra computation \cite{cheng2020decoupling}. To reduce the difficulty of modeling channel-wise topologies, channel-wise topology refinement graph convolution \cite{chen2021channel} models channel-wise topologies by learning a shared topology as a generic prior for all channels and then refining it with channel-specific correlations for each channel.
Generally speaking, the most recent work focuses on designing an adjacency matrix so that the GCN can learn global topology information between skeleton joint points more effectively. However, how to construct an adjacency matrix for interactive action is ignored by existing methods. We present a MTE module to extract mutual adjacency matrix $\textbf{A}_{k+1} \in \mathbb{R}^{2\times C_{k+1}\times N\times N}$ from $\textbf{X}_k$.

\textbf{Feature Generation Block (FGB)}. \label{sec33-1}
For an two-person interactive action, the skeleton sequence $\textbf{X}_k$ can be decomposed as two skeleton sequences, namely, $\textbf{R}_k$ and $\textbf{L}_k$, according to following formulation $\textbf{X}_k = [\textbf{R}_k \:||\: \textbf{L}_k]$, where $||$ denotes concatenation operation, $\textbf{R}_k$ and $\textbf{L}_k$ $\in$ $\mathbb{R}^{C_k\times T_k\times N}$. For one skeleton sequence $\textbf{R}_k$, we calculate the correlation map $\textbf{V}_{\textbf{R}_k}$ via:
$\textbf{V}_{\textbf{R}_k} = \mathcal{N}\Big(\rho\big(\psi(\textbf{R}_k)\big), \rho\big(\phi(\textbf{R}_k)\big)\Big)$,
where $\psi\big(\cdot\big)$ and $\phi\big(\cdot\big)$ are two functions implemented by 1x1 convolution which move the channel dimension of $\textbf{R}_k$ from $C_k$ to $C_k/r$. Therefore $\psi(\textbf{R}_k)$ and $\phi(\textbf{R}_k)$ $\in \mathbb{R}^{C_k/r\times T_k\times N}$, where $r$ is a hyper-parameter. According to the definition of adjacency matrix, we measure the correlation between pairwise skeleton joints. The feature of each skeleton joint belongs to $(\mathbb{R}^{C_k/r\times T_k)\times 1}$. Considering that the temporal axis contains redundant information which 
limits the discriminate ability of skeleton joint features, we use a $\rho\big(\cdot\big)$ function, i.e., average pooling, to reduce the skeleton joint feature to lower feature space $\mathbb{R}^{C_k/r\times 1}$. 
Assume $\textbf{X}, \textbf{Y} \in \mathbb{R}^{C_k/r\times N}$, we define the $ij$-th element of $\mathcal{N}$ as: 
\begin{equation}\label{eq4}
\mathcal{N}_{ij} (\textbf{X},\textbf{Y}) = tanh\left(\textbf{X}[:, i] - \textbf{Y}[:, j]\right),
\end{equation}
where $\mathcal{N}_{ij} (\textbf{X},\textbf{Y}) \in \mathbb{R}^{C_k/r\times 1}$, and $tanh$ is an activation function which normalizes the output to the scope of -1 to 1.
Similar as $\textbf{V}_{\textbf{R}_k}$, we calculate the correlation map $\textbf{V}_{\textbf{L}_k}$ by:
$\textbf{V}_{\textbf{L}_k} = \mathcal{N}\Big(\rho\big(\psi(\textbf{L}_k)\big), \rho\big(\phi(\textbf{L}_k)\big)\Big)$ for $\textbf{L}_k$.

Both $\textbf{V}_{\textbf{R}_k}$ and $\textbf{V}_{\textbf{L}_k}$ belong to $\mathbb{R}^{C_k/r\times N\times N}$, and they denote intra-correlation inside of each skeleton sequence. For each skeleton joint, the importance of this skeleton joint increases when other joints have a strong correlation with it. 
For interactive action, the importance of each skeleton joint is determined by the skeleton joints of both persons. Therefore, we model the inter-correlation between $\textbf{R}_k$ and $\textbf{L}_k$ as$\textbf{V}_{\textbf{R}_k, \textbf{L}_k} = \mathcal{N}\Big(\frac{\rho\big(\psi(\textbf{R}_k)\big)+\rho\big(\phi(\textbf{R}_k)\big)}{2}, \frac{\rho\big(\psi(\textbf{L}_k)\big)+\rho\big(\phi(\textbf{L}_k)\big)}{2}\Big)$,
where $\frac{\rho\big(\psi(\textbf{R}_k)\big)+\rho\big(\phi(\textbf{R}_k)\big)}{2}$ and $\frac{\rho\big(\psi(\textbf{L}_k)\big)+\rho\big(\phi(\textbf{L}_k)\big)}{2}$ mean the average feature of $\textbf{R}_k$ and $\textbf{L}_k$, and $\textbf{V}_{\textbf{R}_k, \textbf{L}_k}$ also belongs to $\mathbb{R}^{C_k/r\times N\times N}$. Since $\textbf{V}_{\textbf{R}_k, \textbf{L}_k}$ is a symmetrical definition, then $\textbf{V}_{\textbf{R}_k, \textbf{L}_k}$ is equal to $\textbf{V}_{\textbf{L}_k, \textbf{R}_k}$.

\textbf{Feature Fusion Block (FFB)}.  \label{sec33-2}
A simple way to jointly model the intra-correlation and the inter-correlation is to fuse them by adding them together. This way ignores that intra-correlation and inter-correlation show different significance in determining the importance of each skeleton joint. Therefore, we modify the adding operation by:
\begin{equation}\label{eq8}
    \tilde{\textbf{V}}_{\textbf{R}_k} = \mathcal{R}(\textbf{V}_{\textbf{R}_k}, \textbf{V}_{\textbf{R}_k, \textbf{L}_k}, \beta) = \textbf{V}_{\textbf{R}_k} + \beta \textbf{V}_{\textbf{R}_k, \textbf{L}_k},
\end{equation}
where $\beta$ is a parameter, whose actual value is learned through backward propagation, and it is multiplied with the inter-correlation $\textbf{V}_{\textbf{R}_k, \textbf{L}_k}$ before adding to the intra-correlation $\textbf{V}_{\textbf{R}_k}$. Compared with $\textbf{V}_{\textbf{R}_k}$ which only contains intra-correlation of skeleton joints from $\textbf{R}_k$, the generated $\tilde{\textbf{V}}_{\textbf{R}_k}$ fuses both intra-correlation of skeleton joints from $\textbf{R}_k$ and inter-correlation of skeleton joints between $\textbf{R}_k$ and $\textbf{L}_k$. Following above pipeline, we formulate $\tilde{\textbf{V}}_{\textbf{L}_k}$ as:
\begin{equation}
    \tilde{\textbf{V}}_{\textbf{L}_k} = \mathcal{R}(\textbf{V}_{\textbf{L}_k}, \textbf{V}_{\textbf{R}_k, \textbf{L}_k}, \beta) = \textbf{V}_{\textbf{L}_k} + \beta \textbf{V}_{\textbf{R}_k, \textbf{L}_k}.
\end{equation}
Similar as Eq.(4), $\tilde{\textbf{V}}_{\textbf{L}_k}$ fuses both the intra-correlation of skeleton joints from $\textbf{L}_k$ and inter-correlation of skeleton joints between $\textbf{R}_k$ and $\textbf{L}_k$.


\textbf{MTE.} Till now, we formulate the MTE module upon FGB and FFB.
For skeleton sequence $\textbf{R}_k$, $\tilde{\textbf{V}}_{\textbf{R}_k} \in \mathbb{R}^{C_k/r\times N\times N}$ is a learnable channel-wise adjacency matrix. To increase the correlation modeling power in the channel axis, we further map this adjacency matrix to $\xi{(\tilde{\textbf{V}}_{\textbf{R}_k})}$, whose channel number is $C_{k+1}$. We implement $\xi\big(\cdot\big)$ function as $1\times1$ convolution.
Our adjacency matrix is further reformulated as
$\textbf{A}_{\textbf{R}_k} = \mathcal{F}\big(\xi{(\tilde{\textbf{V}}_{\textbf{R}_k})}, \textbf{A}, \alpha\big) = \xi{(\tilde{\textbf{V}}_{\textbf{R}_k})} \oplus \alpha \textbf{A}$,
where $\textbf{A}$ is a single channel adjacency matrix that is extracted according to the physical connection of skeleton joints, $\alpha$ is a learnable parameter that can be updated through backward propagation, and $\oplus$ function denotes sum operation for two matrices with different dimensions. Since $\xi{(\tilde{\textbf{V}}_{\textbf{R}_k})}$ belongs to $\mathbb{R}^{C_{k+1}\times N\times N}$ and $\alpha\textbf{A}$ belong to $\mathbb{R}^{1\times N\times N}$, we first repeat the channel dimension of $\alpha\textbf{A}$ for $C_{k+1}$ times and then add it to $\xi{(\tilde{\textbf{V}}_{\textbf{R}_k})}$.
Following above pipeline, we formulate $\textbf{A}_{\textbf{L}_k}$ as $\textbf{A}_{\textbf{L}_k} = \mathcal{F}\big(\xi{(\tilde{\textbf{V}}_{\textbf{L}_k})}, \textbf{A}, \alpha\big) = \xi{(\tilde{\textbf{V}}_{\textbf{L}_k})} \oplus \alpha \textbf{A}$. Both $\textbf{A}_{\textbf{R}_k}$ and $\textbf{A}_{\textbf{L}_k}$ belong to $\mathbb{R}^{C_{k+1}\times N\times N}$. Compared with $\tilde{\textbf{V}}_{\textbf{R}_k}$ and $\tilde{\textbf{V}}_{\textbf{L}_k}$ which are simply learned in the training stage, $\textbf{A}_{\textbf{R}_k}$ and $\textbf{A}_{\textbf{L}_k}$ involve both static and learned adjacency matrices, thus show more flexibility in modeling channel-wise typologies. After obtaining $\textbf{A}_{\textbf{R}_k}$ and $\textbf{A}_{\textbf{L}_k}$, we combine them by:
\begin{equation}
    \textbf{A}_{k+1} = [\textbf{A}_{\textbf{R}_k} \:||\: \textbf{A}_{\textbf{L}_k}],
\end{equation}
where $||$ denotes concatenation operation, and the generated $\textbf{A}_{k+1}$ describes the channel-wise typologies of two persons during performing the interactive action. 

\textbf{MFE.} 
The general pipeline of the MFE module is inspired by the MTE module. Given a skeleton sequence $\textbf{X}_k$, we follow above formulation and decompose $\textbf{X}_k$ as two skeleton sequences, namely, $\textbf{R}_k$ and $\textbf{L}_k$, where $\textbf{R}_k$ and $\textbf{L}_k$ $\in \mathbb{R}^{C_k\times T_k\times N}$. 
First, we use 1 $\times$ 1 convolution to first extract local features for both $\textbf{R}_k$ and $\textbf{L}_k$.
Second, we simply use feature averaging to obtain the shared feature.
Third, we use FFB to dynamically inject the shared feature into the individual feature. The output feature of our mutual feature excitation block belongs to $\mathbb{R}^{2\times C_{k+1}\times T_{k}\times N}$.

\section{Experiments}
We conduct experiments on three datasets including Assemble101 dataset \cite{sener2022assembly101} for hand-to-hand interaction recognition, NTU60-Interaction dataset \cite{shahroudy2016ntu} and NTU120-Interaction dataset \cite{liu2019ntu} for human-to-human interaction recognition. We compare our method with LSTM methods, e.g., LSTM-IRN \cite{perez2021interaction}, GCN methods, e.g., ST-GCN \cite{yan2018spatial}, Transformer methods, e.g., IGFormer \cite{pang2022igformer}, and Graph Transformer methods, e.g. 2sKA-AGTN \cite{liu2022graph}.

\subsection{Datasets and Settings}
\textbf{Assemble101} dataset \cite{sener2022assembly101} is a large-scale multi-view video dataset for understanding procedural activities. 
It is annotated with 46286 action segments for the training set and 21900 action segments for the test set.
There are 90 objects and 24 interaction verbs, which form a total of \textbf{\textit{1380 fine-grained actions}}.
Considering the scale of total action segments and the number of action labels, it is the largest and most challenging dataset for hand-to-hand interaction recognition. To ensure a fair comparison with previous methods, we just use the skeleton sequential data of two human hands as inputs.

\textbf{NTU60-Interaction} dataset origins from NTU-RGB+D dataset \cite{shahroudy2016ntu}, which is the most popular skeleton-based action recognition dataset that contains 56,578 samples covering 60 action labels. According to \cite{pang2022igformer}, we use the 11 human interaction classes covering ``A1. punching/slapping other person", ``A2. kicking other person", ``A3. pushing other person", ``A4. pat on back of other person", ``A5. point finger at the other person", ``A6. hugging other person", ``A7. giving something to other person", ``A8. touch other person's pocket", ``A9. handshaking", ``A10. walking towards each other", ``A11. walking apart from each other". 

\textbf{NTU120-Interaction} dataset is built upon the original NTU-RGB+D-120 dataset \cite{liu2019ntu}, which is one of the largest skeleton-based action recognition datasets that contains 113,945 samples covering 120 action labels. According to \cite{pang2022igformer}, we use the 26 human interaction classes. Compared with the NTU60-Interaction dataset, this dataset additionally covers ``A12. hit other person with something", ``A13. wield knife towards other person", ``A14. knock over other person", ``A15. grab other person’s stuff", ``A16. shoot at other person with a gun", ``A17. step on foot", ``A18. high-five", ``A19. cheers and drink", ``A20. carry something with other person", ``A21. take a photo of other person", ``A22. follow other person", ``A23. whisper in other person’s ear", ``A24. exchange things with other person", ``A25. support somebody with hand", ``A26. finger-guessing game". Interaction classes like ``A13. wield knife towards other person", ``A14. knock over other person" and ``A15. grab other person’s stuff" share similar interactive movements that make it difficult to classify these actions. Following \cite{pang2022igformer}, we use cross subject (X-Sub) protocol and cross view (X-View) protocol for the NTU-60-Interaction dataset and use cross subject (X-Sub) protocol and cross setup (X-Set) protocol for NTU-120-Interaction dataset.

\textbf{Settings}.
All experiments are conducted on one Tesla V100 GPU with PyTorch. We adopt the SGD algorithm \cite{bottou2012stochastic} with a Nesterov momentum of 0.9 and weight decay of 0.0004 as the optimizer. To make the training procedure more stable, the first 5 epochs are used for warm-up. The initial learning rate is set to 0.01 and is divided by 10 at the 35th and 55th epochs. The training process is terminated at the 65th epoch, and the batch size is set to 32 for both datasets. To ensure fair comparisons, we adopt the same preprocessing steps as in previous methods. For the NTU60-Interaction and NTU120-Interaction datasets, we follow the procedures outlined in \cite{zhang2020semantics,li2018co}, removing the global movements from each skeleton sequence by subtracting the coordinates of each skeleton joint from the coordinates of the second joint in the first frame. For the Assemble101 dataset, we use the original data provided by \cite{sener2022assembly101} without applying any additional preprocessing steps. Skeleton sequences with different temporal frames are resized to 64 frames. We set $r$ to 8 as default value for all datasets.

\begin{table}[]
\centering
\resizebox{18em}{!}
{\begin{tabular}{cccc}
\hline
\textbf{Method}  & \textbf{Acc. (\%)} \\ \hline
2s-AGCN \cite{shi2019two}              & 22.2 \\
IGFormer \cite{pang2022igformer}               & 22.5                \\
\textbf{Ours}         & \textbf{27.2}   $^{\uparrow{4.7}}$      \\ \hline
\end{tabular}}
\caption{Comparison on Assemble 101 dataset}\label{tb1}
\end{table}

\begin{table}[t]
\centering
\resizebox{30em}{!}
{\begin{tabular}{ccccc}
\hline
\textbf{Method}  & \textbf{X-View (\%)}  & \textbf{X-Sub (\%)} \\ \hline
ST-GCN \cite{yan2018spatial}                  & 87.1     & 83.3                     \\
ST-LSTM \cite{liu2016spatio}               & 87.3           & 83.0                   \\
GCA \cite{liu2017global}                      & 89.0     & 85.9                     \\
2s-GCA \cite{liu2017skeleton}                 & 89.9      & 87.2                   \\
ViT \cite{dosovitskiy2020image}              & 92.5      & 89.7                    \\
AS-GCN \cite{li2019actional}                   & 93.0     & 89.3                   \\
LSTM-IRN \cite{perez2021interaction}        & 93.5      & 90.5                    \\
2sKA-AGTN \cite{liu2022graph}            & 96.1    & 90.4 \\
IGFormer \cite{pang2022igformer}          & 96.5       & 
93.6                    \\
\textbf{Ours}          & \textbf{98.2}   $^{\uparrow{1.7}}$                     & \textbf{95.5}  $^{\uparrow{1.9}}$    \\ 
\hline
\end{tabular}}
\caption{Comparison on NTU60-Interaction dataset}\label{tb2}
\end{table}

\begin{table}[t]
\centering
\resizebox{28em}{!}
{\begin{tabular}{ccccc}
\hline
\textbf{Method}  & \textbf{X-Set (\%)} & \textbf{X-Sub (\%)}  \\ \hline
ST-LSTM \cite{liu2016spatio}                  & 66.6             & 63.0               \\
2s-GCA \cite{liu2017skeleton}              & 73.3        & 73.0                   \\
GCA \cite{liu2017global}                      & 73.7     & 70.6                   \\
ST-GCN \cite{yan2018spatial}                 & 76.1     & 78.9                     \\
LSTM-IRN \cite{perez2021interaction}           & 79.6      & 77.7                 \\
ViT \cite{dosovitskiy2020image}           & 82.5   & 81.5                      \\
AS-GCN \cite{li2019actional}              & 83.7      & 82.9                   \\
IGFormer \cite{pang2022igformer}             & 86.5      &85.4                    \\
2sKA-AGTN \cite{liu2022graph}         &  88.2       &  86.7 \\
\textbf{Ours}       & \textbf{90.0}   $^{\uparrow{1.8}}$           & \textbf{90.0}   $^{\uparrow{3.3}}$   \\ \hline
\end{tabular}}
\caption{Comparison on NTU120-Interaction dataset}\label{tb3}
\end{table}

\subsection{Comparison with the State-of-the-Art}
We compare our method with state-of-the-art methods on Assemble101 dataset (see Table \ref{tb1}), NTU60-Interaction dataset (see Table \ref{tb2}), and NTU120-Interaction dataset (see Table \ref{tb3}). Generally, ours achieves the best performances on three datasets.
Specifically, for the hand-to-hand interaction task, IGFormer achieves an accuracy of 22.5\% on the Assemble101 dataset. The low performance is caused by two reasons. First, the Assemble101 dataset is challenging as it contains 1380 action types, which is far larger than other action recognition datasets. Second, many actions share similar hand motion patterns that are difficult to distinguish.
Compared with IGFormer, we achieve an accuracy of 27.2\%, which is 4.7\% higher than IGFormer. Considering that this dataset is rather challenging, our improvement is non-trivial.

For the human-to-human interaction task, IGFormer achieves an accuracy of 93.6\% using the X-Sub protocol and achieves an accuracy of 96.5\% using the X-View protocol on the NTU60-Interaction dataset. Our method outperforms IGFormer using both protocols. 
Since the performances on the NTU60-Interaction dataset are nearly saturated, we use a larger NTU120-Interaction dataset for comparison. 2sKA-AGTN, the state-of-the-art Graph Transformer-based method, achieves an accuracy of 88.2\% using the X-Set protocol and achieves an accuracy of 86.7\% using the X-Sub protocol. Compared with 2sKA-AGTN, we achieve an accuracy of 90.0\% using the X-Set protocol and achieves an accuracy of 90.0\% using the X-Sub protocol, which outperforms 2sKA-AGTN by 1.8\% and 3.3\%. These improvements verify that our me-GC can encode additional rich mutual information which benefits the identification of interactive actions. IGFormer, the state-of-the-art transformer-based method, achieves an accuracy of 86.5\% using the X-Set protocol and achieves an accuracy of 85.4\% using the X-Sub protocol. Our method achieves improvements of 3.5\% and 4.6\% over IGFormer. Compared with transformer architecture, we infer the reason that GCN with mutual learning is a better fit for modeling the spatial relationships of skeleton-based interactive motion patterns.

\begin{table}[]
\centering
\resizebox{28em}{!}
{\begin{tabular}{ccccc}
\hline
\textbf{Baseline} & \textbf{MTE} & \textbf{MFE}  & \textbf{TC} & \textbf{Acc. (\%)} \\ \hline
\checkmark                 &             &              &              &      84.2              \\
\checkmark                 &    \checkmark           &           &              &    86.8 $^{\uparrow{2.6}}$               \\
\checkmark                 &            &     \checkmark           &            &    86.8 $^{\uparrow{2.6}}$               \\ \hline
\checkmark                 &            &              &   \checkmark           &    87.8                 \\
\checkmark                 & \checkmark           & \checkmark            & \checkmark            & 90.0  $^{\uparrow{2.2}}$                 \\ \hline
\end{tabular}}
\caption{Ablation study on different blocks}\label{tb4}
\end{table}

\subsection{Ablation Study}
Considering the popularity and the scale of the dataset for interactive action recognition, we adopt the NTU120-Interaction dataset and X-Sub protocol for ablation study.
We set up a GCN called ``\textbf{\textit{Baseline}}", which stacks ten graph convolution layers. To implement each graph convolution, we use FGB to generate adjacency matrices and use 1 $\times$ 1 convolution for extracting local features.
We also present a strong baseline called ``\textbf{\textit{Baseline + TC}}" by additionally involving temporal convolution operation after each graph convolution layer.
Both Baseline and Baseline + TC follow the split-and-fusion pipeline (see Fig. \ref{fig1} (b)).


\textbf{Effect of MTE \& MFE. }
Table \ref{tb4} evaluates the effect of our proposed MTE module and MFE module.
Compared with the Baseline, our MTE achieves an accuracy of 86.8\%, which outperforms the Baseline by 2.6\%. Meanwhile, our MFE achieves an accuracy of 86.8\%, which also outperforms the Baseline by 2.6\%.
These results verify the effect of mutual learning for pure GCN.
Baseline + TC achieves an accuracy of 87.8\%.
By additionally applying MTE and MFE, we achieve an accuracy of 90.0\%, which further improves Baseline + TC by 2.2\%.
This improvement verifies that \textbf{\textit{mutual learning and individual spatial-temporal encoding show complementary property to each other}}.

\begin{figure*}[htbp]
	\centering
	\subfigure[Baseline]{\includegraphics[width=.48\linewidth]{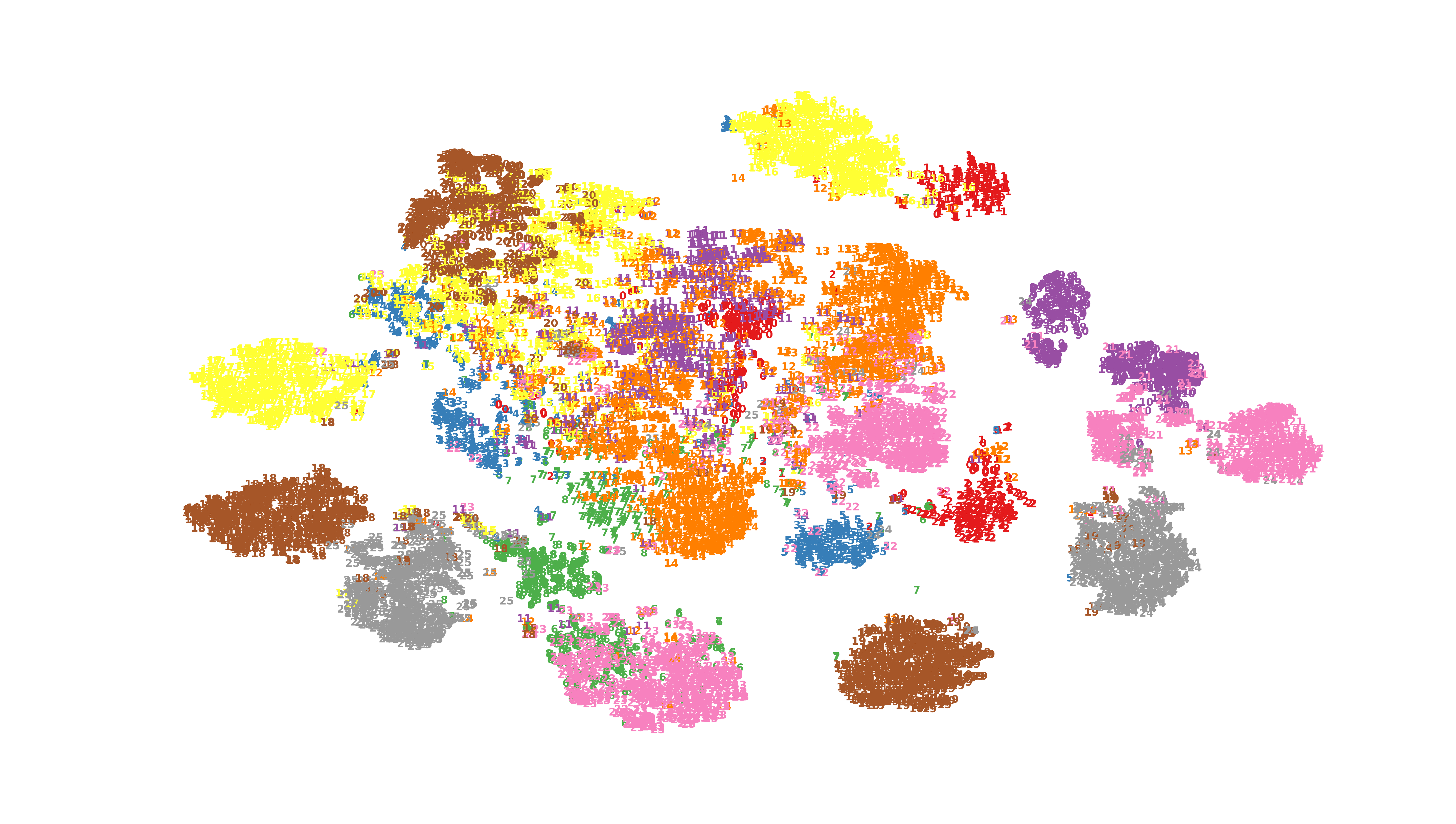}}\hspace{5pt}
	\subfigure[MTE only]{\includegraphics[width=.48\linewidth]{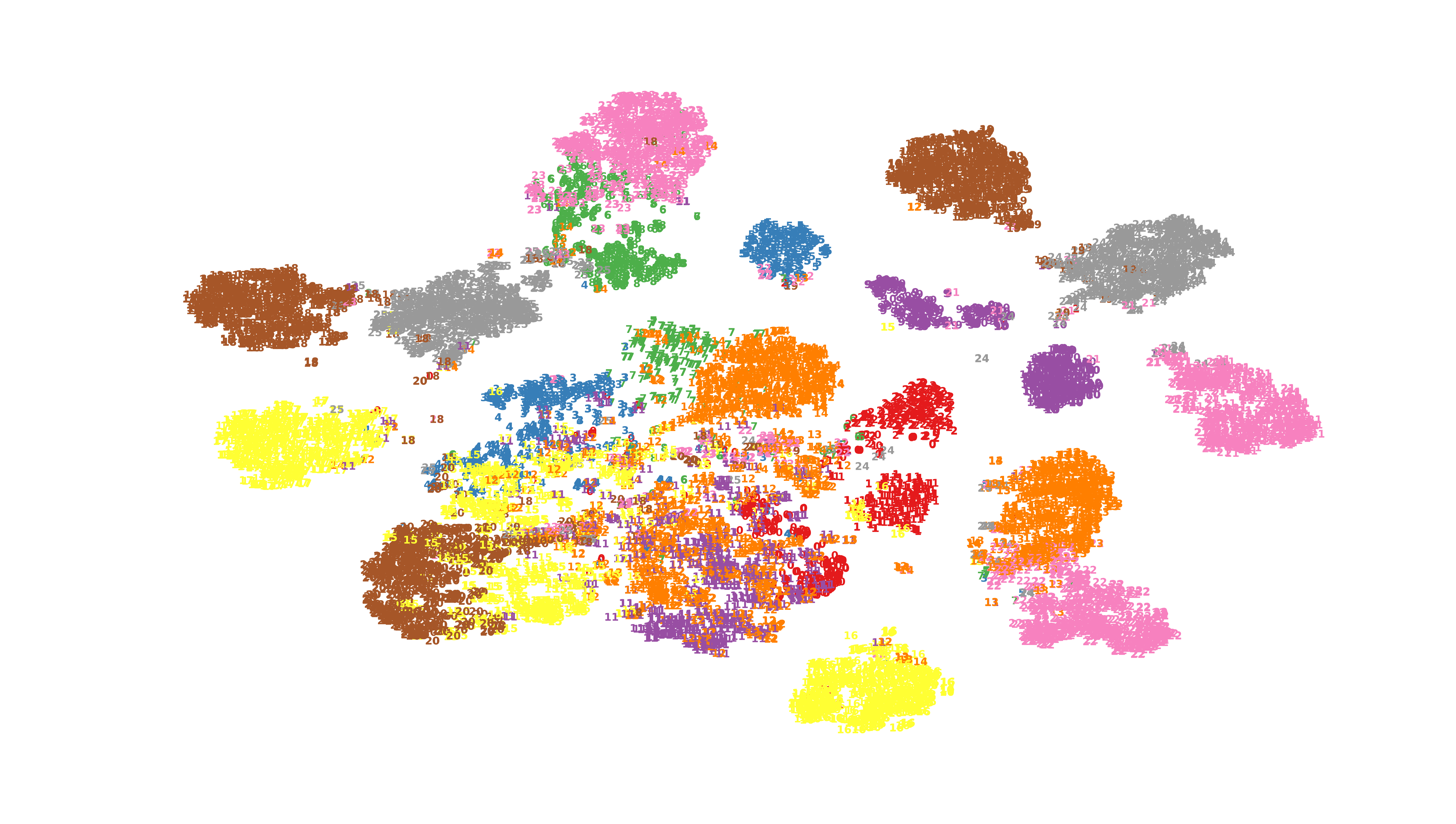}}\\
	\subfigure[MFE only]{\includegraphics[width=.48\linewidth]{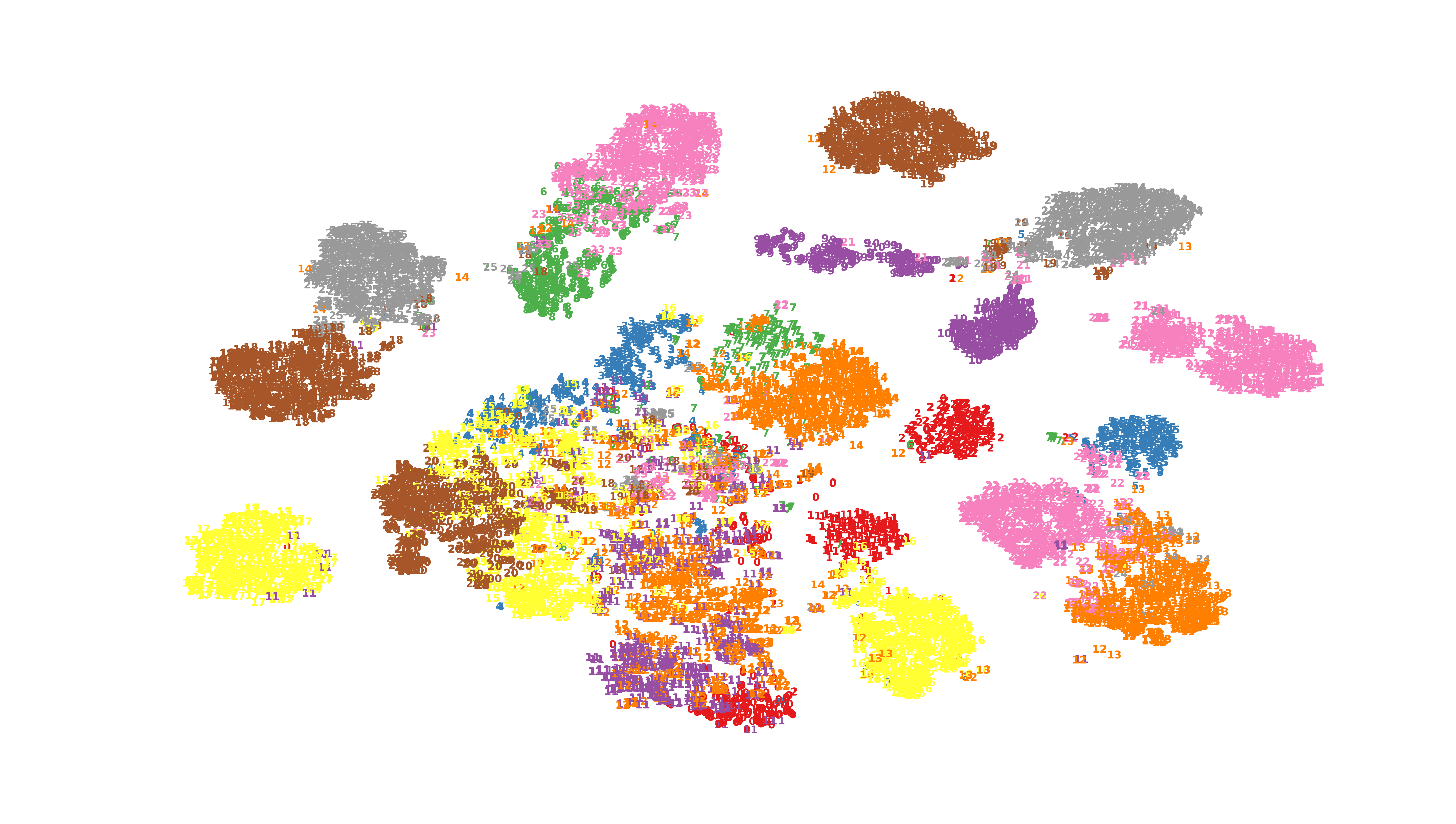}}\hspace{5pt}
	\subfigure[Ours]{\includegraphics[width=.48\linewidth]{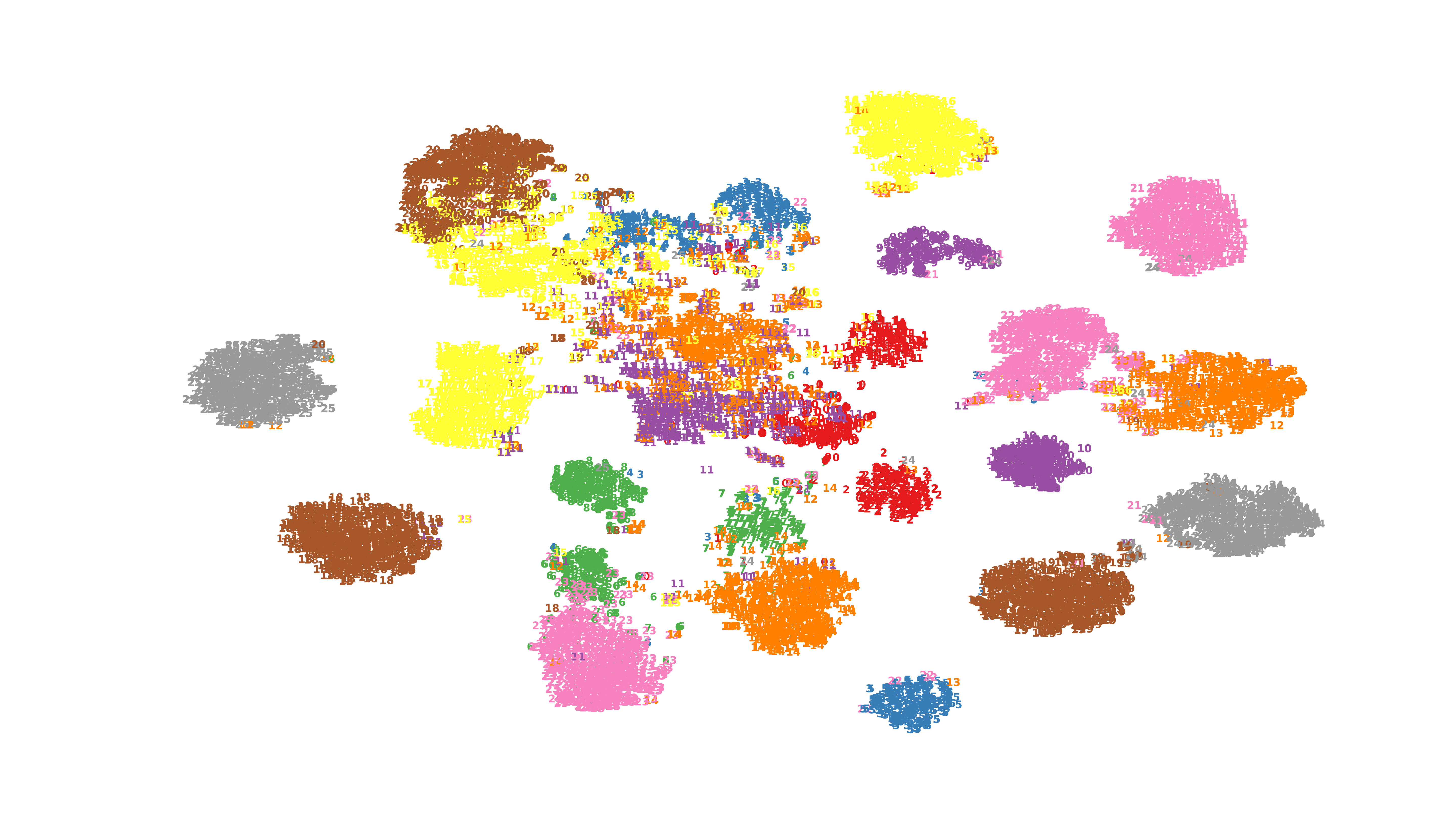}}
	\caption{t-SNE \cite{van2008visualizing} visualization of skeleton sequence representation on the test set of the NTU120-Interaction dataset learned by Baseline method and different variants of our model: MTE only, MFE only, and Ours. Compared with the Baseline, ours learns more distinctive representations to differentiate similar interactive actions. Noting that we use both shape and color to denote different actions.}\label{fig4}
\end{figure*}

\begin{table}[]
\centering
\resizebox{28em}{!}
{\begin{tabular}{ccccccccccc}
\hline
\multicolumn{10}{c}{\textbf{Layer}}                                                                                              & \multirow{2}{*}{\textbf{Acc. (\%)}} \\ \cline{1-10}
\textbf{1} & \textbf{2} & \textbf{3} & \textbf{4} & \textbf{5} & \textbf{6} & \textbf{7} & \textbf{8} & \textbf{9} & \textbf{10} &                                     \\ \hline
&&&&&&&&&& 87.8\\
\checkmark          & \checkmark          & \checkmark          & \checkmark          &            &            &            &            &            &             &             89.8                \\
           &            &            & \checkmark          & \checkmark          & \checkmark          & \checkmark          &            &            &             &    88.6   \\
           &            &            &            &            &            & \checkmark          & \checkmark          & \checkmark          & \checkmark           &     89.2   \\
\checkmark          & \checkmark          & \checkmark          & \checkmark          & \checkmark          & \checkmark          & \checkmark          & \checkmark          & \checkmark          & \checkmark           &   90.0                                 \\ \hline
\end{tabular}}
\vspace{-0.5em}
\caption{Ablation study on different layers}\label{tb5}
\vspace{-0.em}
\end{table}

\begin{table}[t]
\centering
\resizebox{24em}{!}
{\begin{tabular}{ccc}
\hline
\textbf{Initialization}  & \textbf{MTE (\%)}              & \textbf{MFE (\%)} \\ \hline
0.0                  & 90.0              & 90.0       \\ 
0.2                  & 90.0             & 89.8     \\ 
0.4                  & 89.4             & 90.1     \\
0.6                  & 89.0             & 89.9     \\ 
0.8                  & 89.0             & 89.9     \\ 
1.0                  & 89.5             & 90.1     \\ \hline
\end{tabular}}
\vspace{-0.5em}
\caption{Ablation study on parameter initialization}\label{tb6}
\end{table}

\begin{table}[t]
\centering
\resizebox{28em}{!}
{\begin{tabular}{cccccc}
\hline
\textbf{Num. of Layers} & 8    & 9    & 10   & 11   & 12   \\ \hline
\textbf{Acc. (\%)}      & 89.8 & 89.9 & 90.0 & 90.0 & 89.3 \\ \hline
\end{tabular}}
\vspace{-0.5em}
\caption{Ablation study on the number of layers}\label{tb7}
\vspace{-0.em}
\end{table}

\begin{table}[t]
\centering
\resizebox{16em}{!}
{\begin{tabular}{cc}
\hline
\textbf{Method} & \textbf{Acc. (\%)} \\ \hline
Early Fusion    &   82.1                 \\
Late Fusion     &   87.8                 \\
Ours            &  90.0    \\ \hline
\end{tabular}}
\vspace{-0.5em}
\caption{Comparison with alternative methods}\label{tb8}
\vspace{-1em}
\end{table}

\textbf{Effect of me-GC Layer. }
Each me-GC layer contains an MTE module and an MFE module.
Table \ref{tb5} shows the ablation study on selecting different layers to use the me-GC layer (marked with \checkmark) instead of the original GC layer (no mark).
When all ten layers use GC, the model is Baseline + TC, which achieves an accuracy of 87.8\%.
We test applying our me-GC layers on the early stage (from 1st layer to 4th layer), middle stage (from 4th layer to 7th layer), and late stage (from 7th to 10th layer) of Baseline + TC. Compared with Baseline + TC, ours achieve at least 0.8\% improvement. By applying our me-GC layers on all ten layers, we achieve an accuracy of 90.0\%, which outperforms Baseline + TC by 2.2\%. These results consistently verify the effect of our me-GC layer over the original GC layer.

\begin{figure*}[t]
  \centering
   \includegraphics[width=0.95\linewidth]{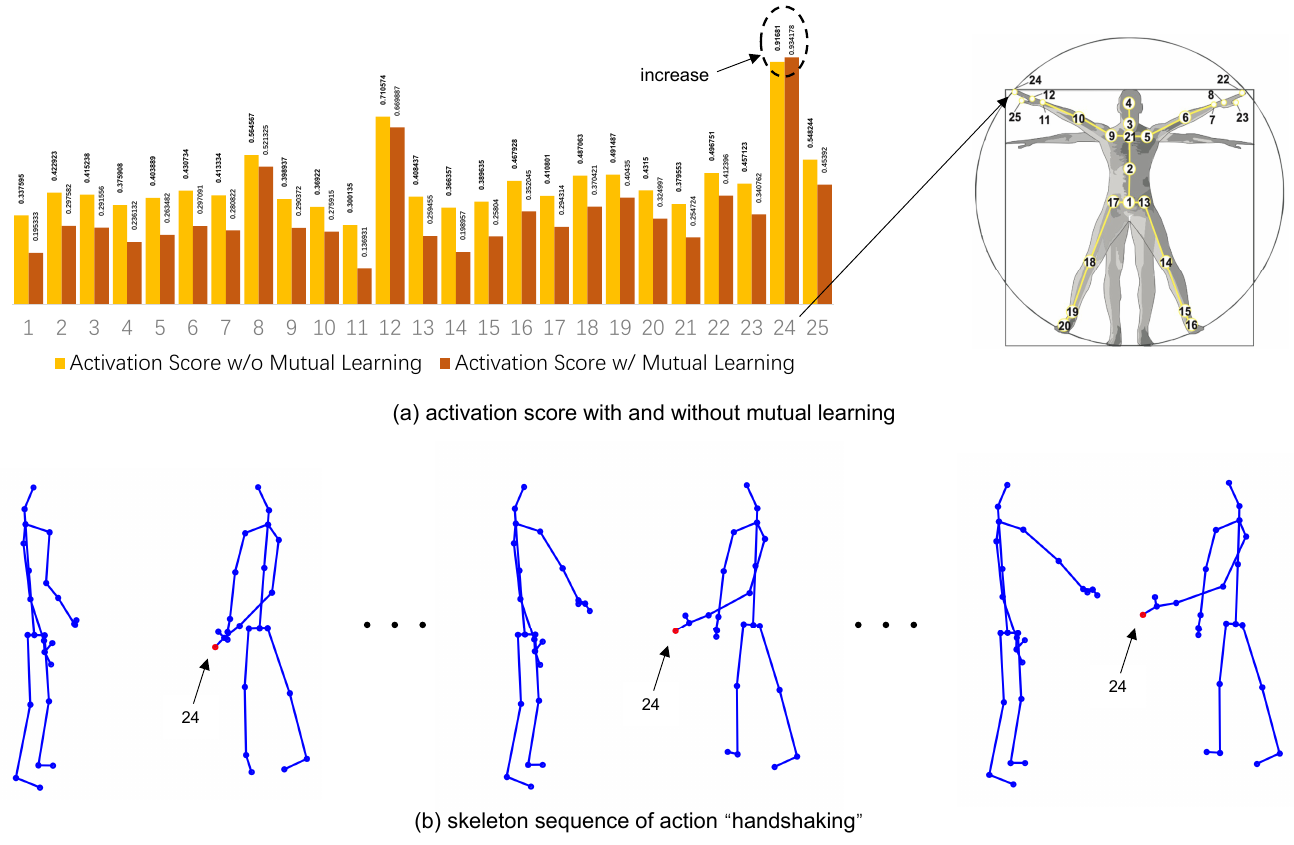}
   \vspace{-0.5em}
   \caption{Effect of mutual learning on the activation score per joint for one entity participating in an ``shaking hands" action. We observe that mutual learning increases the activation score of the critical joint which shows a strong correlation with the action label.}
   \vspace{-0.5em}
   \label{fig5}
\end{figure*}

\begin{figure}[t]
  \centering
   \includegraphics[width=0.96\linewidth]{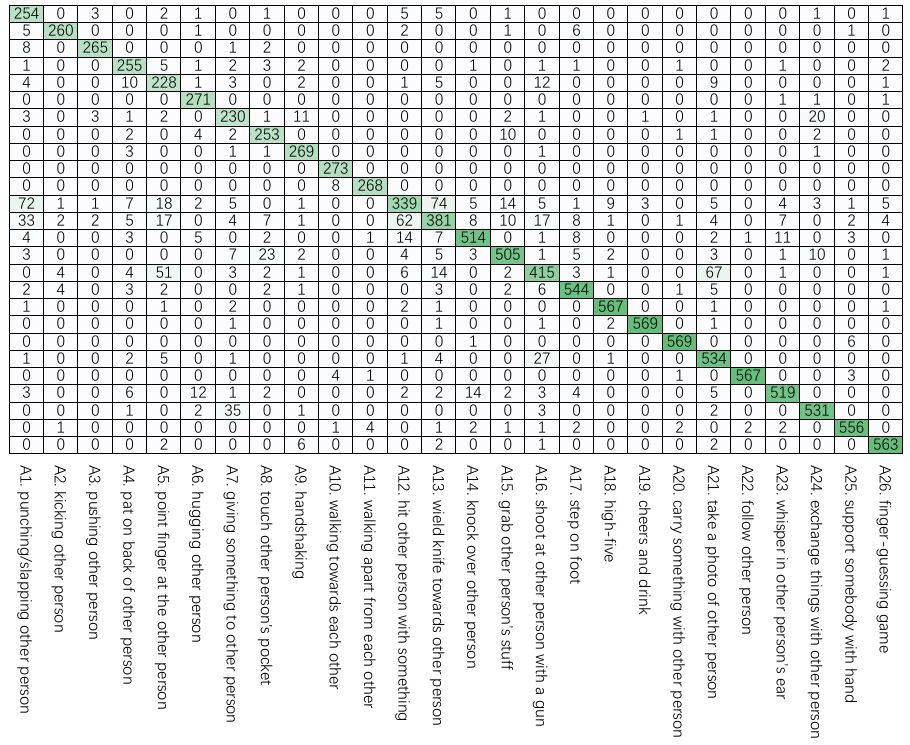}
   \vspace{-1pt}
   \caption{Confusion matrix on NTU120-Interaction dataset using cross subject protocol}
   \label{conf1}
   \vspace{-2pt}
\end{figure}

\begin{figure}[t]
  \centering
   \includegraphics[width=0.96\linewidth]{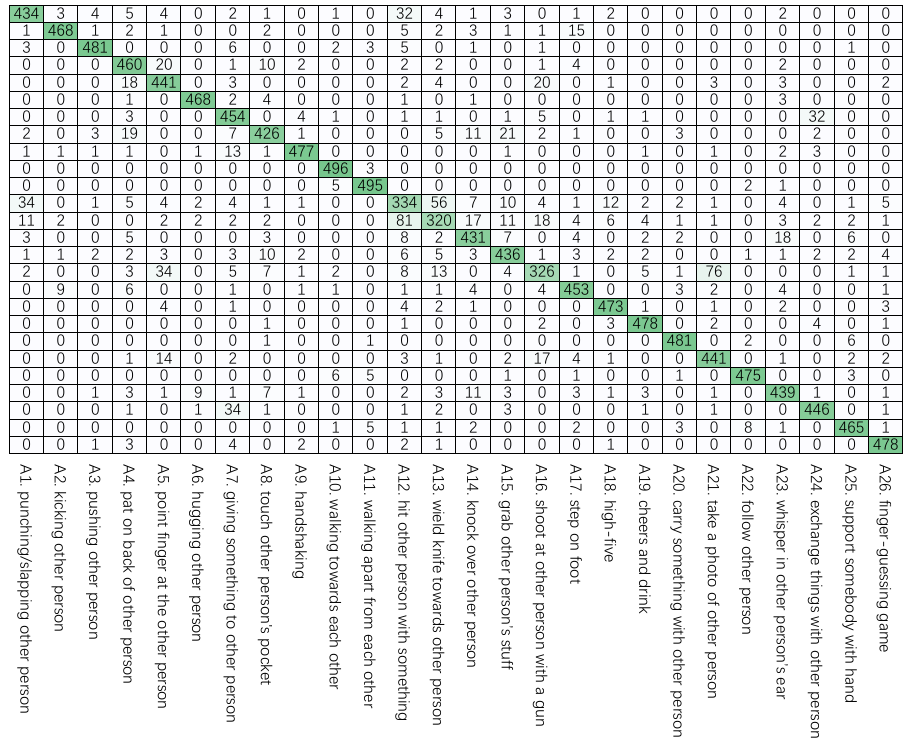}
   \vspace{-1pt}
   \caption{Confusion matrix on NTU120-Interaction dataset using cross setup protocol}
   \label{conf2}
   \vspace{-2pt}
\end{figure}

\textbf{Effect of Initialization. }
For the MTE, parameter $\beta$ determines the fusion of shared adjacency matrix extracted from pairwise skeleton sequences and individual adjacency matrix of each skeleton sequence. Also, parameter $\beta$ determines the fusion of shared features and individual features. To facilitate parameter selection on different datasets, we use a learnable parameter that can be optimized by backpropagation. Table \ref{tb6} evaluates the initialization of parameter $\beta$ that varies from 0.0 to 1.0 at an interval of 0.2. We observe that our method is not sensitive to initialization. For simplicity, we set the initial value of parameter $\beta$ to zero, which achieves competitive results for MTE and MFE.

\textbf{Number of me-GC layers. }
Table \ref{tb7} evaluates the number of me-GC layers that varies from 8 to 12. We achieve the best performances when using 10 layers and 11 layers. We choose the default value as 10 for all three datasets, considering the Cuda memory and computational cost.

\textbf{Comparison with alternative methods. }
To fuse information from pairwise skeleton sequences, we present me-GCN to learn mutual relationships in cascade me-GC layers, which achieves an accuracy of 90.0\%. We provide two simple alternative methods called ``Early Fusion" and ``Late Fusion" for learning mutual interaction. Specifically, ``Early Fusion" fuses pairwise skeleton sequences at the data level, and ``Late Fusion" fuses deep features extracted from pairwise skeleton sequences before the inference layer. As shown in Table \ref{tb8}, the alternative Early Fusion method achieves an accuracy of 82.1\%, which is 7.9\% lower than our method. We infer the reason that \textbf{\textit{early fusion destroys the motion patterns of individual entities}}. Another alternative Late Fusion method achieves an accuracy of 87.8\%, which is 2.2\% lower than ours. The reason is that \textbf{\textit{late fusion ignores the mutual learning of low-level features}}, which play a more significant role in encoding mutual semantic cues among interactive body parts.

\subsection{Visualization}
\textbf{Feature representation}. Fig. \ref{fig4} shows the t-SNE \cite{van2008visualizing} visualization of skeleton sequence representation.
We project the skeleton sequence representation extracted from the NTU120-Interaction test dataset to the 2-D dimension using t-SNE.
We show 26 types of interactive actions, which are denoted by different shapes and colored by continuous colors.
As can be seen, learned representations from variants of our model, i.e., MTE only and MFE only, can be grouped slightly better than the Baseline. Moreover, by combining MTE and MFE, our method can better differentiate skeleton sequence representations than the Baseline. These results verify the discriminant ability of our method.

\textbf{Activation score}. 
Fig. \ref{fig5} shows the activation score of each skeleton joint for one entity that comes from an ``shaking hands" action of the NTU120-Interaction dataset. The activation score reflects the correlation between the skeleton joint and the action label. To calculate the activation score, we aggregate and normalize the last me-GC layer's multi-channel adjacency matrices as a 1 $\times$ 25 vector, where each value denotes the activation score of one joint. Inside the MTE module, we compare the activation score before the FFB block and after the FFB block and find that the activation score of the 24th joint increases, meanwhile the activation score of other joints decreases. As the joint means human hand and is strongly related to the action ``shaking hands", this phenomenon verifies that \textbf{\textit{mutual learning benefits encoding distinctive body parts}}.

\textbf{Confusion matrix}.
To illustrate the performances of our method on each action, we show confusion matrices of our method on the NTU120-Interaction dataset using cross subject protocol (see Fig. \ref{conf1}) and cross setup protocol (see Fig. \ref{conf2}). We observe that our method achieves high accuracy on various actions. For example, the accuracy of action ``A25. support somebody with hand" and the accuracy of action ``A26. finger-guessing game" are more than 95\% using both protocols.

\section{Conclusion and Future Work}
We investigate the skeleton-based interactive action recognition task and presents a mutual excitation graph convolutional network (me-GCN) by stacking mutual excitation graph convolution (me-GC) layers.
Specifically, our proposed me-GC layer adopts a mutual topology excitation (MTE) module for topology modeling and a mutual feature excitation (MFE) module for feature extraction.
Beyond the standard graph convolution layer, our me-GC layer can learn mutual semantic relationships between interactive body parts, thus facilitating the encoding of critical interactive motion patterns.
Compared with GCN and Transformer-based methods, we achieve state-of-the-art performances on two challenging human-to-human interaction datasets and one large-scale hand-to-hand interaction dataset.
Future work focus on combining our method with recent Transformer-based methods like \cite{do2025skateformer}.

\bibliographystyle{IEEEtran}
\bibliography{egbib.bib}

\begin{thebibliography}{10}
\providecommand{\url}[1]{#1}
\csname url@samestyle\endcsname
\providecommand{\newblock}{\relax}
\providecommand{\bibinfo}[2]{#2}
\providecommand{\BIBentrySTDinterwordspacing}{\spaceskip=0pt\relax}
\providecommand{\BIBentryALTinterwordstretchfactor}{4}
\providecommand{\BIBentryALTinterwordspacing}{\spaceskip=\fontdimen2\font plus
\BIBentryALTinterwordstretchfactor\fontdimen3\font minus \fontdimen4\font\relax}
\providecommand{\BIBforeignlanguage}[2]{{%
\expandafter\ifx\csname l@#1\endcsname\relax
\typeout{** WARNING: IEEEtran.bst: No hyphenation pattern has been}%
\typeout{** loaded for the language `#1'. Using the pattern for}%
\typeout{** the default language instead.}%
\else
\language=\csname l@#1\endcsname
\fi
#2}}
\providecommand{\BIBdecl}{\relax}
\BIBdecl

\bibitem{naser2022privacy}
A.~Naser, A.~Lotfi, M.~D. Mwanje, and J.~Zhong, ``Privacy-preserving, thermal vision with human in the loop fall detection alert system,'' \emph{IEEE Transactions on Human-Machine Systems}, vol.~53, no.~1, pp. 164--175, 2022.

\bibitem{wang2022predicting}
W.~Wang, J.~Li, Y.~Li, and X.~Dong, ``Predicting activities of daily living for the coming time period in smart homes,'' \emph{IEEE Transactions on Human-Machine Systems}, vol.~53, no.~1, pp. 228--238, 2022.

\bibitem{hadikhani2023novel}
P.~Hadikhani, D.~T.~C. Lai, and W.-H. Ong, ``A novel skeleton-based human activity discovery using particle swarm optimization with gaussian mutation,'' \emph{IEEE Transactions on Human-Machine Systems}, 2023.

\bibitem{taghanaki2023self}
S.~R. Taghanaki, M.~Rainbow, and A.~Etemad, ``Self-supervised human activity recognition with localized time-frequency contrastive representation learning,'' \emph{IEEE Transactions on Human-Machine Systems}, 2023.

\bibitem{baruah2023intent}
M.~Baruah, B.~Banerjee, and A.~K. Nagar, ``Intent prediction in human--human interactions,'' \emph{IEEE Transactions on Human-Machine Systems}, vol.~53, no.~2, pp. 458--463, 2023.

\bibitem{liu2022generalized}
M.~Liu, F.~Meng, and Y.~Liang, ``Generalized pose decoupled network for unsupervised 3d skeleton sequence-based action representation learning,'' \emph{Cyborg and Bionic Systems}, 2022.

\bibitem{liu2023temporal}
J.~Liu, X.~Wang, C.~Wang, Y.~Gao, and M.~Liu, ``Temporal decoupling graph convolutional network for skeleton-based gesture recognition,'' \emph{IEEE Transactions on Multimedia}, 2023.

\bibitem{wang2024dynamic}
X.~Wang, W.~Zhang, C.~Wang, Y.~Gao, and M.~Liu, ``Dynamic dense graph convolutional network for skeleton-based human motion prediction.'' \emph{IEEE Transactions on Image Processing}, 2024.

\bibitem{zhang2024facial}
Y.~Zhang, X.~Xu, Y.~Zhao, Y.~Wen, Z.~Tang, and M.~Liu, ``Facial prior guided micro-expression generation,'' \emph{IEEE Transactions on Image Processing}, 2024.

\bibitem{wang2023global}
Y.~Wang, H.~Kang, D.~Wu, W.~Yang, and L.~Zhang, ``Global and local spatio-temporal encoder for 3d human pose estimation,'' \emph{IEEE Transactions on Multimedia}, 2023.

\bibitem{tu2023dtcm}
Z.~Tu, Y.~Liu, Y.~Zhang, Q.~Mu, and J.~Yuan, ``Dtcm: Joint optimization of dark enhancement and action recognition in videos,'' \emph{IEEE Transactions on Image Processing}, 2023.

\bibitem{tu2023consistent}
Z.~Tu, Z.~Huang, Y.~Chen, D.~Kang, L.~Bao, B.~Yang, and J.~Yuan, ``Consistent 3d hand reconstruction in video via self-supervised learning,'' \emph{IEEE Transactions on Pattern Analysis and Machine Intelligence (TPAMI)}, vol.~45, no.~8, pp. 9469--9485, 2023.

\bibitem{gao2023dual}
Q.~Gao, Z.~Deng, Z.~Ju, and T.~Zhang, ``Dual-hand motion capture by using biological inspiration for bionic bimanual robot teleoperation,'' \emph{Cyborg and Bionic Systems (CBS)}, vol.~4, p. 0052, 2023.

\bibitem{zhang2016rgb}
J.~Zhang, W.~Li, P.~O. Ogunbona, P.~Wang, and C.~Tang, ``Rgb-d-based action recognition datasets: A survey,'' \emph{Pattern Recognition}, vol.~60, pp. 86--105, 2016.

\bibitem{ahmad2021graph}
T.~Ahmad, L.~Jin, X.~Zhang, S.~Lai, G.~Tang, and L.~Lin, ``Graph convolutional neural network for human action recognition: a comprehensive survey,'' \emph{IEEE Transactions on Artificial Intelligence}, vol.~2, no.~2, pp. 128--145, 2021.

\bibitem{pareek2021survey}
P.~Pareek and A.~Thakkar, ``A survey on video-based human action recognition: recent updates, datasets, challenges, and applications,'' \emph{Artificial Intelligence Review}, vol.~54, no.~3, pp. 2259--2322, 2021.

\bibitem{kong2022human}
Y.~Kong and Y.~Fu, ``Human action recognition and prediction: A survey,'' \emph{International Journal of Computer Vision}, vol. 130, no.~5, pp. 1366--1401, 2022.

\bibitem{zhang2012microsoft}
Z.~Zhang, ``Microsoft kinect sensor and its effect,'' \emph{IEEE Multimedia}, vol.~19, no.~2, pp. 4--10, 2012.

\bibitem{keselman2017intel}
L.~Keselman, J.~Iselin~Woodfill, A.~Grunnet-Jepsen, and A.~Bhowmik, ``Intel realsense stereoscopic depth cameras,'' in \emph{IEEE Conference on Computer Vision and Pattern Recognition Workshops (CVPRW)}, 2017, pp. 1--10.

\bibitem{liu2017enhanced}
M.~Liu, H.~Liu, and C.~Chen, ``Enhanced skeleton visualization for view invariant human action recognition,'' \emph{Pattern Recognition}, vol.~68, pp. 346--362, 2017.

\bibitem{sun2022human}
Z.~Sun, Q.~Ke, H.~Rahmani, M.~Bennamoun, G.~Wang, and J.~Liu, ``Human action recognition from various data modalities: A review,'' \emph{IEEE Transactions on Pattern Analysis and Machine Intelligence}, 2022.

\bibitem{perez2021interaction}
M.~Perez, J.~Liu, and A.~C. Kot, ``Interaction relational network for mutual action recognition,'' \emph{IEEE Transactions on Multimedia}, vol.~24, pp. 366--376, 2021.

\bibitem{pang2022igformer}
Y.~Pang, Q.~Ke, H.~Rahmani, J.~Bailey, and J.~Liu, ``Igformer: Interaction graph transformer for skeleton-based human interaction recognition,'' in \emph{European Conference on Computer Vision (ECCV)}, 2022, pp. 605--622.

\bibitem{schuster1997bidirectional}
M.~Schuster and K.~K. Paliwal, ``Bidirectional recurrent neural networks,'' \emph{IEEE Transactions on Signal Processing}, vol.~45, no.~11, pp. 2673--2681, 1997.

\bibitem{vaswani2017attention}
A.~Vaswani, N.~Shazeer, N.~Parmar, J.~Uszkoreit, L.~Jones, A.~N. Gomez, {\L}.~Kaiser, and I.~Polosukhin, ``Attention is all you need,'' \emph{Advances in Neural Information Processing Systems (NeurIPS)}, vol.~30, 2017.

\bibitem{yan2018spatial}
S.~Yan, Y.~Xiong, and D.~Lin, ``Spatial temporal graph convolutional networks for skeleton-based action recognition,'' in \emph{AAAI conference on Artificial Intelligence (AAAI)}, 2018, pp. 7444--7452.

\bibitem{cheng2020decoupling}
K.~Cheng, Y.~Zhang, C.~Cao, L.~Shi, J.~Cheng, and H.~Lu, ``Decoupling gcn with dropgraph module for skeleton-based action recognition,'' in \emph{European Conference on Computer Vision (ECCV)}, 2020, pp. 536--553.

\bibitem{li2019actional}
M.~Li, S.~Chen, X.~Chen, Y.~Zhang, Y.~Wang, and Q.~Tian, ``Actional-structural graph convolutional networks for skeleton-based action recognition,'' in \emph{IEEE/CVF Conference on Computer Vision and Pattern Recognition (CVPR)}, 2019, pp. 3595--3603.

\bibitem{zhang2012spatio}
Y.~Zhang, X.~Liu, M.-C. Chang, W.~Ge, and T.~Chen, ``Spatio-temporal phrases for activity recognition,'' in \emph{European Conference on Computer Vision (ECCV)}, 2012, pp. 707--721.

\bibitem{ji2014interactive}
Y.~Ji, G.~Ye, and H.~Cheng, ``Interactive body part contrast mining for human interaction recognition,'' in \emph{IEEE International Conference on Multimedia and Expo Workshops (ICMEW)}, 2014, pp. 1--6.

\bibitem{ji2015learning}
Y.~Ji, H.~Cheng, Y.~Zheng, and H.~Li, ``Learning contrastive feature distribution model for interaction recognition,'' \emph{Journal of Visual Communication and Image Representation}, vol.~33, pp. 340--349, 2015.

\bibitem{wu2017recognition}
H.~Wu, J.~Shao, X.~Xu, Y.~Ji, F.~Shen, and H.~T. Shen, ``Recognition and detection of two-person interactive actions using automatically selected skeleton features,'' \emph{IEEE Transactions on Human-Machine Systems}, vol.~48, no.~3, pp. 304--310, 2017.

\bibitem{du2015hierarchical}
Y.~Du, W.~Wang, and L.~Wang, ``Hierarchical recurrent neural network for skeleton based action recognition,'' in \emph{IEEE Conference on Computer Vision and Pattern Recognition (CVPR)}, 2015, pp. 1110--1118.

\bibitem{liu2016spatio}
J.~Liu, A.~Shahroudy, D.~Xu, and G.~Wang, ``Spatio-temporal lstm with trust gates for 3d human action recognition,'' in \emph{European Conference on Computer Vision (ECCV)}, 2016, pp. 816--833.

\bibitem{plizzari2021skeleton}
C.~Plizzari, M.~Cannici, and M.~Matteucci, ``Skeleton-based action recognition via spatial and temporal transformer networks,'' \emph{Computer Vision and Image Understanding}, vol. 208, p. 103219, 2021.

\bibitem{du2015skeleton}
Y.~Du, Y.~Fu, and L.~Wang, ``Skeleton based action recognition with convolutional neural network,'' in \emph{IAPR Asian Conference on Pattern Recognition (ACPR)}, 2015, pp. 579--583.

\bibitem{ke2017new}
Q.~Ke, M.~Bennamoun, S.~An, F.~Sohel, and F.~Boussaid, ``A new representation of skeleton sequences for 3d action recognition,'' in \emph{IEEE Conference on Computer Vision and Pattern Recognition (CVPR)}, 2017, pp. 3288--3297.

\bibitem{peng2020learning}
W.~Peng, X.~Hong, H.~Chen, and G.~Zhao, ``Learning graph convolutional network for skeleton-based human action recognition by neural searching,'' in \emph{AAAI Conference on Artificial Intelligence (AAAI)}, vol.~34, no.~03, 2020, pp. 2669--2676.

\bibitem{shi2019two}
L.~Shi, Y.~Zhang, J.~Cheng, and H.~Lu, ``Two-stream adaptive graph convolutional networks for skeleton-based action recognition,'' in \emph{IEEE/CVF Conference on Computer Vision and Pattern Recognition (CVPR)}, 2019, pp. 12\,026--12\,035.

\bibitem{chen2021channel}
Y.~Chen, Z.~Zhang, C.~Yuan, B.~Li, Y.~Deng, and W.~Hu, ``Channel-wise topology refinement graph convolution for skeleton-based action recognition,'' in \emph{IEEE International Conference on Computer Vision (ICCV)}, 2021, pp. 13\,359--13\,368.

\bibitem{chi2022infogcn}
H.-g. Chi, M.~H. Ha, S.~Chi, S.~W. Lee, Q.~Huang, and K.~Ramani, ``Infogcn: Representation learning for human skeleton-based action recognition,'' in \emph{IEEE/CVF Conference on Computer Vision and Pattern Recognition (CVPR)}, 2022, pp. 20\,186--20\,196.

\bibitem{paszke2019pytorch}
A.~Paszke, S.~Gross, F.~Massa, A.~Lerer, J.~Bradbury, G.~Chanan, T.~Killeen, Z.~Lin, N.~Gimelshein, L.~Antiga \emph{et~al.}, ``Pytorch: An imperative style, high-performance deep learning library,'' \emph{Advances in Neural Information Processing Systems (NeurIPS)}, vol.~32, 2019.

\bibitem{srivastava2014dropout}
N.~Srivastava, G.~Hinton, A.~Krizhevsky, I.~Sutskever, and R.~Salakhutdinov, ``Dropout: a simple way to prevent neural networks from overfitting,'' \emph{Journal of Machine Learning Research}, vol.~15, no.~1, pp. 1929--1958, 2014.

\bibitem{zhang2019graph}
S.~Zhang, H.~Tong, J.~Xu, and R.~Maciejewski, ``Graph convolutional networks: a comprehensive review,'' \emph{Computational Social Networks}, vol.~6, no.~1, pp. 1--23, 2019.

\bibitem{huang2020spatio}
Z.~Huang, X.~Shen, X.~Tian, H.~Li, J.~Huang, and X.-S. Hua, ``Spatio-temporal inception graph convolutional networks for skeleton-based action recognition,'' in \emph{ACM International Conference on Multimedia (ACM MM)}, 2020, pp. 2122--2130.

\bibitem{ye2020dynamic}
F.~Ye, S.~Pu, Q.~Zhong, C.~Li, D.~Xie, and H.~Tang, ``Dynamic gcn: Context-enriched topology learning for skeleton-based action recognition,'' in \emph{ACM International Conference on Multimedia (ACM MM)}, 2020, pp. 55--63.

\bibitem{sener2022assembly101}
F.~Sener, D.~Chatterjee, D.~Shelepov, K.~He, D.~Singhania, R.~Wang, and A.~Yao, ``Assembly101: A large-scale multi-view video dataset for understanding procedural activities,'' in \emph{IEEE/CVF Conference on Computer Vision and Pattern Recognition (CVPR)}, 2022, pp. 21\,096--21\,106.

\bibitem{shahroudy2016ntu}
A.~Shahroudy, J.~Liu, T.-T. Ng, and G.~Wang, ``Ntu rgb+d: A large scale dataset for 3d human activity analysis,'' in \emph{IEEE Conference on Computer Vision and Pattern Recognition (CVPR)}, 2016, pp. 1010--1019.

\bibitem{liu2019ntu}
J.~Liu, A.~Shahroudy, M.~Perez, G.~Wang, L.-Y. Duan, and A.~C. Kot, ``Ntu rgb+d 120: A large-scale benchmark for 3d human activity understanding,'' \emph{IEEE Transactions on Pattern Analysis and Machine Intelligence}, vol.~42, no.~10, pp. 2684--2701, 2019.

\bibitem{liu2022graph}
Y.~Liu, H.~Zhang, D.~Xu, and K.~He, ``Graph transformer network with temporal kernel attention for skeleton-based action recognition,'' \emph{Knowledge-Based Systems}, vol. 240, p. 108146, 2022.

\bibitem{bottou2012stochastic}
L.~Bottou, ``Stochastic gradient descent tricks,'' \emph{Neural Networks: Tricks of the Trade: Second Edition}, pp. 421--436, 2012.

\bibitem{zhang2020semantics}
P.~Zhang, C.~Lan, W.~Zeng, J.~Xing, J.~Xue, and N.~Zheng, ``Semantics-guided neural networks for efficient skeleton-based human action recognition,'' in \emph{IEEE/CVF Conference on Computer Vision and Pattern Recognition (CVPR)}, 2020, pp. 1112--1121.

\bibitem{li2018co}
C.~Li, Q.~Zhong, D.~Xie, and S.~Pu, ``Co-occurrence feature learning from skeleton data for action recognition and detection with hierarchical aggregation,'' in \emph{International Joint Conference on Artificial Intelligence (IJCAI)}, 2018, pp. 786--792.

\bibitem{liu2017global}
J.~Liu, G.~Wang, P.~Hu, L.-Y. Duan, and A.~C. Kot, ``Global context-aware attention lstm networks for 3d action recognition,'' in \emph{IEEE Conference on Computer Vision and Pattern Recognition (CVPR)}, 2017, pp. 1647--1656.

\bibitem{liu2017skeleton}
J.~Liu, G.~Wang, L.-Y. Duan, K.~Abdiyeva, and A.~C. Kot, ``Skeleton-based human action recognition with global context-aware attention lstm networks,'' \emph{IEEE Transactions on Image Processing}, vol.~27, no.~4, pp. 1586--1599, 2017.

\bibitem{dosovitskiy2020image}
A.~Dosovitskiy, L.~Beyer, A.~Kolesnikov, D.~Weissenborn, X.~Zhai, T.~Unterthiner, M.~Dehghani, M.~Minderer, G.~Heigold, S.~Gelly \emph{et~al.}, ``An image is worth 16x16 words: Transformers for image recognition at scale,'' in \emph{International Conference on Learning Representations (ICLR)}, 2021.

\bibitem{van2008visualizing}
L.~Van~der Maaten and G.~Hinton, ``Visualizing data using t-sne.'' \emph{Journal of Machine Learning Research}, vol.~9, no.~11, 2008.

\bibitem{do2025skateformer}
J.~Do and M.~Kim, ``Skateformer: skeletal-temporal transformer for human action recognition,'' in \emph{European Conference on Computer Vision (ECCV)}, 2025, pp. 401--420.

\end{thebibliography}

\end{document}